\pdfoutput=1
\PassOptionsToPackage{dvipsnames}{xcolor}
\documentclass[11pt]{article}

\usepackage[]{ACL2023}

\usepackage{times}
\usepackage{latexsym}

\usepackage[T1]{fontenc}

\usepackage[utf8]{inputenc}

\usepackage{microtype}

\usepackage{inconsolata}
\usepackage{latexsym}
\usepackage{float}
\usepackage{arydshln}
\usepackage{booktabs}
\usepackage{amssymb}
\usepackage{amsmath, calc}
\usepackage{multirow}
\usepackage{graphicx}
\usepackage{makecell}
\usepackage{subfigure}
\usepackage{algorithm}
\usepackage{setspace}
\usepackage{enumitem}

\usepackage[noend]{algcompatible}
\usepackage[normalem]{ulem}
\useunder{\uline}{\ul}{}

\makeatletter
\def\adl@drawiv#1#2#3{%
        \hskip.5\tabcolsep
        \xleaders#3{#2.5\@tempdimb #1{1}#2.5\@tempdimb}%
                #2\z@ plus1fil minus1fil\relax
        \hskip.5\tabcolsep}
\newcommand{\cdashlinelr}[1]{%
  \noalign{\vskip\aboverulesep
           \global\let\@dashdrawstore\adl@draw
           \global\let\adl@draw\adl@drawiv}
  \cdashline{#1}
  \noalign{\global\let\adl@draw\@dashdrawstore
           \vskip\belowrulesep}}
\makeatother

\title{Large Language Models are Zero-Shot Relation Extractors\\ with Aligned Instruction Task}
\title{Aligning Instruction Tasks Unlocks Large Language Models\\ as Zero-Shot Relation Extractors}

\author{Kai Zhang \quad Bernal Jim\'{e}nez Guti\'{e}rrez \quad Yu Su \\[1pt]
        The Ohio State University\\
        {\small \texttt{\{zhang.13253, jimenezgutierrez.1, su.809\}@osu.edu}}
}

\begin{document}
\maketitle
\begin{abstract}

Recent work has shown that fine-tuning large language models (LLMs) on large-scale instruction-following datasets substantially improves their performance on a wide range of NLP tasks, especially in the zero-shot setting.
However, even advanced instruction-tuned LLMs still fail to outperform small LMs on relation extraction (RE), a fundamental information extraction task.
We hypothesize that instruction-tuning has been unable to elicit strong RE capabilities in LLMs due to RE's low incidence in instruction-tuning datasets, making up less than 1\% of all tasks \cite{Wang2022SuperInstructions}.
To address this limitation, we propose QA4RE, a framework that aligns RE with question answering (QA), a predominant task in instruction-tuning datasets.
Comprehensive zero-shot RE experiments over four datasets with two series of instruction-tuned LLMs (six LLMs in total) demonstrate that our QA4RE framework consistently improves LLM performance, strongly verifying our hypothesis and enabling LLMs to outperform strong zero-shot baselines by a large margin.
Additionally, we provide thorough experiments and discussions to show the robustness, few-shot effectiveness, and strong transferability of our QA4RE framework.
This work illustrates a promising way of adapting LLMs to challenging and underrepresented tasks by aligning these tasks with more common instruction-tuning tasks like QA.\footnote{Code and data are available at \href{https://github.com/OSU-NLP-Group/QA4RE}{https://github.com/OSU-NLP-Group/QA4RE}.
}
\end{abstract}

\section{Introduction}

\begin{figure}[!t]
 \centering
 \small
 \includegraphics[width=\columnwidth]{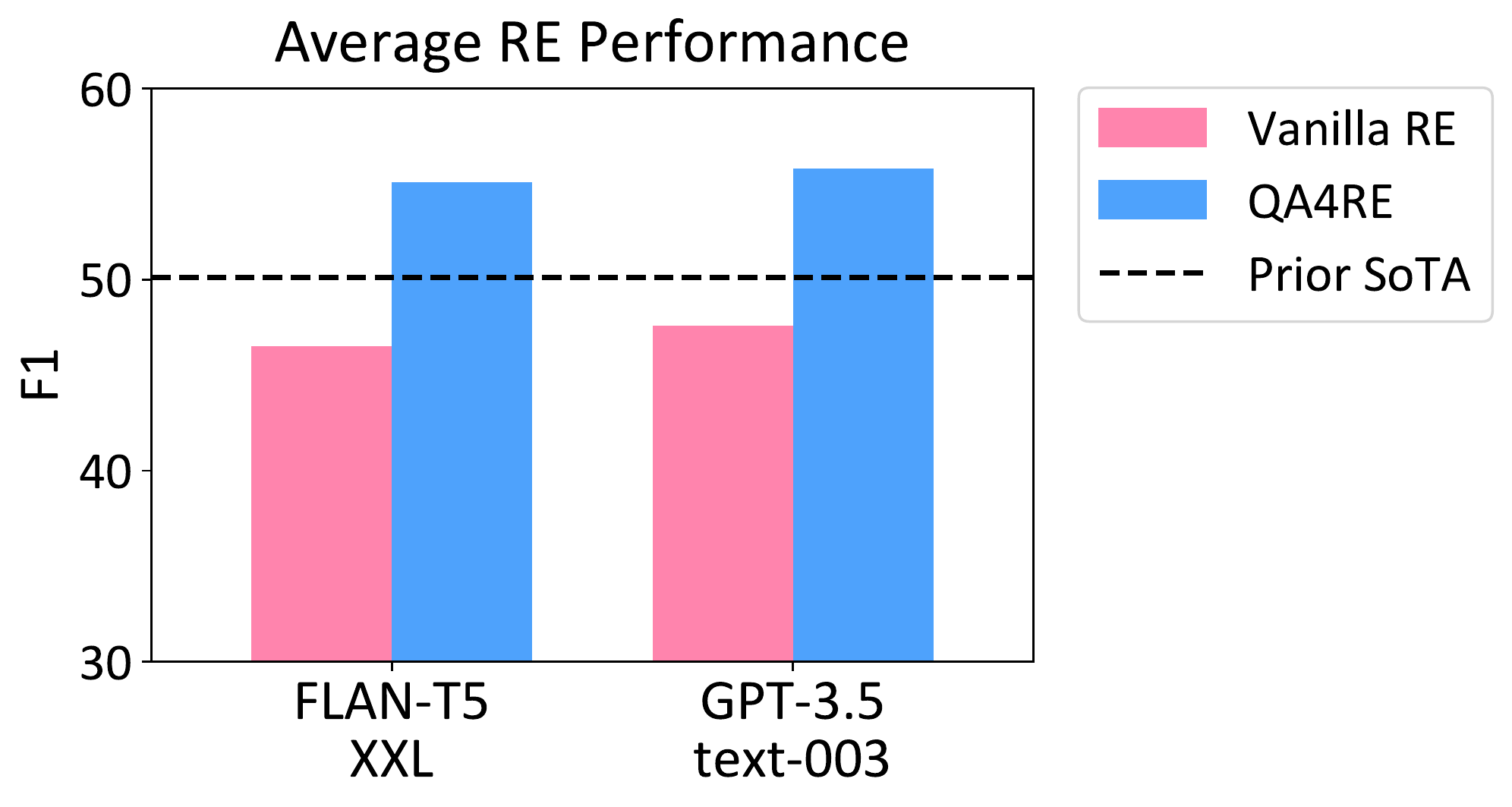}
\caption{Main finding: Strong instruction-tuned LLMs underperform prior zero-shot RE methods using the standard (vanilla) RE formulation. Our QA4RE framework enables models in two sets of instruction-tuned LLMs (FLAN-T5 and GPT-3.5) to surpass the prior SoTA on 4 RE datasets by a large margin. Results are averaged over 4 RE datasets. We omit the word `davinci' from the GPT-3.5 model displayed for brevity.}
        \label{fig:hook-figure}
\vspace{-5pt}
\end{figure}

Large language models (LLMs)~\cite{Brown2020GPT3, Chowdhery2022PaLM, Zhang2022OPT} have been shown to achieve impressive performance on many NLP tasks. 
Using the in-context learning paradigm, without any parameter updating, LLMs are able to achieve comparable performance with small language models (LMs) fine-tuned on thousands of examples ~\cite{Liu2022KATE, Min2022Channel, Liang2022HolisticEO}.\footnote{We regard LMs with less than 1B params as small.}
%
More recently, fine-tuning LLMs on datasets containing thousands of downstream tasks transformed into an instruction following format (i.e., \textit{instruction-tuning}) has been shown to improve LLMs considerably across the board, especially in zero-shot setting~\cite{Iyer2022OPT-IML, Ouyang2022InstructGPT, ChungFlanT5}. 

\begin{figure*}[!t]
\small
 \centering
 \includegraphics[width=\textwidth]{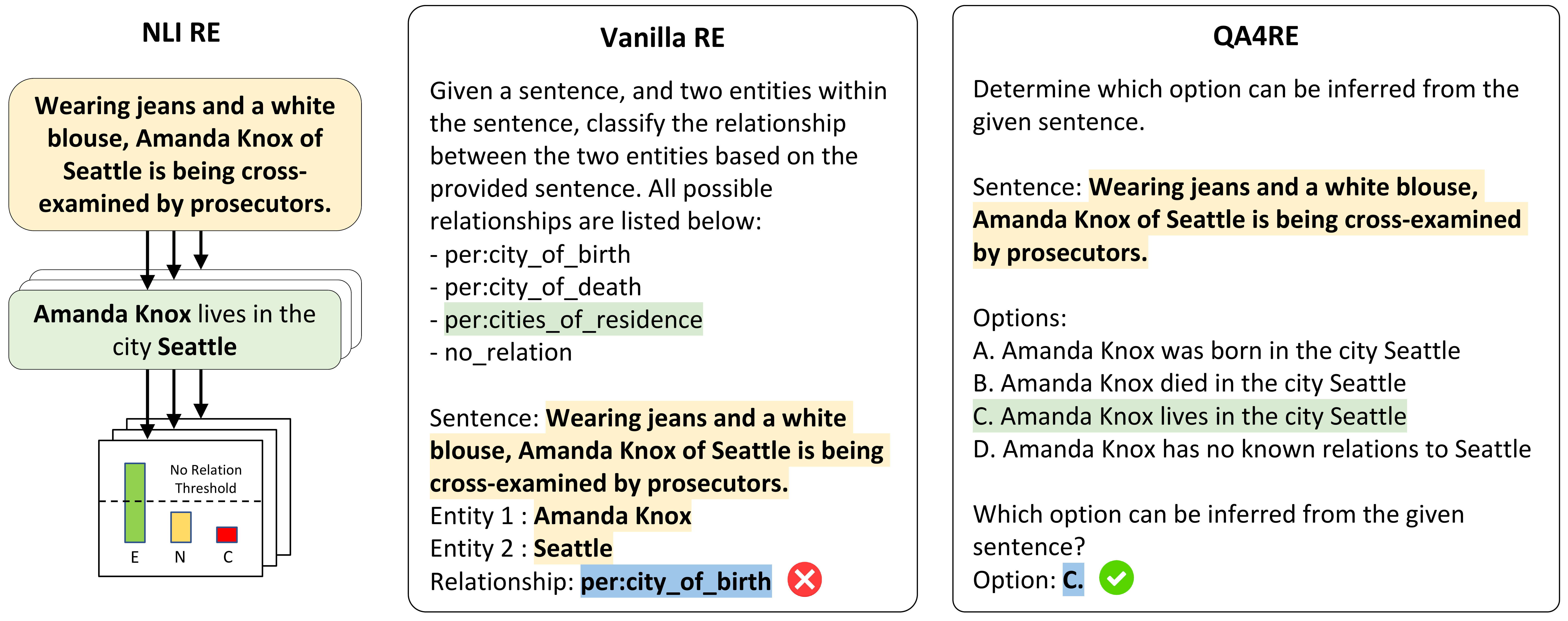}
\caption{This figure shows a schematic of the SoTA NLI zero-shot framework in which each sentence must be compared with each relation template (left), the vanilla formulation for prompting GPT-3 for RE as done in \citet{Gutierrez2022GPT3BioIE} (center) and our multiple-choice QA setting, in which each relation is transformed into a template and GPT-3 is expected to predict only a single letter (right).}
\label{fig:intro}
\vspace{-5pt}
\end{figure*}

We examine the capability of LLMs in identifying the relationship between entities in a sentence, i.e., relation extraction (RE), a fundamental task in information extraction.
Recent work~\cite{Gutierrez2022GPT3BioIE} has found that LLMs underperform fine-tuned small LMs for RE in the biomedical domain. 
Our results on general domain RE in Fig.~\ref{fig:hook-figure} reveal that even two of the most advanced instruction-tuned LLMs, FLAN-T5 XXL~\cite{ChungFlanT5} and text-davinci-003 \cite{Ouyang2022InstructGPT}, fail to outperform the state-of-the-art (SoTA) zero-shot RE method based on small LMs~\cite{Sainz2021NLI}.
We hypothesize that the limited relation extraction capability of instruction-tuned LLMs could be a byproduct of the low incidence of RE tasks in instruction-tuning datasets~\cite{Ouyang2022InstructGPT, Sanh2022T0, ChungFlanT5, Wang2022SuperInstructions}.\footnote{RE-like tasks are <0.5\% of the largest available instruction dataset~\cite{Wang2022SuperInstructions}; see Appendix~\ref{sed:appendix-instruction-dataset-portion} for details.}
To address the low incidence issue, we propose the QA4RE framework, which aligns RE with multiple-choice question answering (QA), a task that appears much more frequently in most instruction-tuning datasets---around $12$-$15$\% of all the tasks in both \citet{Wang2022SuperInstructions} and \citet{Ouyang2022InstructGPT}. 
Specifically, by casting the input sentence as a question and possible relation types as multiple-choice options (Fig.~\ref{fig:intro}), LLMs are able to perform RE by selecting the option representing the correct relation type.

Thorough evaluations on four real-world relation extraction datasets and six instruction-tuned models from two different series (OpenAI GPT-3.5 and FLAN-T5~\cite{ChungFlanT5}) show that QA4RE brings significant gains over the standard RE formulation on, validating its effectiveness and our hypothesis concerning the low incidence of RE.
More specifically, our framework enables text-davinci-003 and FLAN-T5-XXLarge to achieve an average of $8.2$\% and $8.6$\% absolute improvements in F1, respectively.
For the first time, an LLM is able to outperform prior small-LM-based SoTA in the zero-shot setting by a large margin.
In-depth analyses further demonstrate the robustness and few-shot effectiveness of QA4RE. 
More importantly, our framework has been proven to be effectively transferable on instruction-tuned models with various sizes, ranging from 80M to 175B.
Our contributions are summarized as follows:

\noindent
\textbf{(1)}
We systematically investigate instruction-tuned LLMs on four real-world relation extraction datasets and note that their limited performance on RE might stem from the low incidence of RE tasks in instruction-tuning datasets.

\noindent
\textbf{(2)}
We reformulate RE as multiple-choice QA in an effort to appropriately leverage QA's much higher prevalence in instruction-tuning datasets and achieve significant improvements on six recent instruction-tuned LLMs, significantly outperforming previous SoTA zero-shot RE methods based on small LM for the first time.

\noindent
\textbf{(3)}
In addition, we demonstrate our QA4RE method's robustness to diverse prompt designs as well as its promising results in the few-shot setting. 

\noindent
\textbf{(4)}
Finally, we show the effectiveness of QA4RE framework is transferable and consistent on various instruction-tuned models with different sizes from 80M to 175B.
Our study illustrates the potential of aligning infrequent and challenging tasks with frequent instruction-tuning tasks and can guide others in exploring this direction.

\section{Related Work}\label{sec:related_work}

\paragraph{Instruction Tuning.} Large language models originally obtained impressive zero and few-shot performance by leveraging self-supervised next token prediction at massive scales. 
More recently, supervised fine-tuning on a large number of downstream tasks has been shown to improve LLM accuracy, robustness, fairness, and generalization to unseen tasks~\cite{Ouyang2022InstructGPT, Iyer2022OPT-IML, Wei2022FLAN, ChungFlanT5, Sanh2022T0}. 
Several strategies have been developed to align LLMs to human instructions including Reinforcement Learning from Human Feedback (RLHF) \cite{Ouyang2022InstructGPT} as well as the more standard language modeling objective, used to fine-tune LLMs on a wide range of tasks reformulated as instruction following tasks \cite{Iyer2022OPT-IML, Wei2022FLAN, ChungFlanT5, Sanh2022T0}.

\paragraph{Eliciting LLM Abilities.}
The high cost and increasingly private nature of LLM pre-training make it quite challenging to conclusively determine how different pre-training techniques bring about different LLM capabilities.
Many factors involved in pre-training such as simple self-supervised scaling, code or multi-lingual text pre-training \cite{Chowdhery2022PaLM, Chen2021Codex, ChungFlanT5} as well as the distinct versions of instruction-tuning mentioned above \cite{Ouyang2022InstructGPT, Iyer2022OPT-IML, Wei2022FLAN, ChungFlanT5}, can interact in a wide variety of ways to unleash the abilities LLMs display.
Nonetheless, \citet{fu2022gptroadmap} hypothesize that the use of code during pre-training seems to improve an LM's reasoning ability, evidenced by the improved ability to leverage Chain-of-Thought prompting \cite{Wei2022CoT} by models trained partially on code such as PaLM \cite{Chowdhery2022PaLM}, code-davinci-002~\cite{Chen2021Codex}, and text-davinci-002/003~\cite{Ouyang2022InstructGPT}, compared to text-only models like text-davinci-001 and OPT-175B~\cite{Zhang2022OPT}. 
Additionally, instruction-tuning on a large set of tasks has been shown to improve generalization to unseen tasks, reduce the need for few-shot examples and improve accuracy and robustness across many language tasks \cite{Ouyang2022InstructGPT, Iyer2022OPT-IML, ChungFlanT5}.

\paragraph{Low-Resource Relation Extraction.} Several reformulations of standard RE have enabled small LMs to achieve fairly strong performance in the zero and few-shot settings. 
~\citet{Sainz2021NLI} utilize small LMs fine-tuned on natural language inference (NLI) datasets to perform zero-shot RE by selecting the entity-filled relation template which is mostly entailed by the test sentence. 
~\citet{Lu2022SuRE} frame RE as a summarization task and leverage generative models to summarize the relation between target entities in the sentence.
Other low-resource RE methods augment prompt-tuning by using logical rules to create complex prompts from sub-prompts \cite{Han2022PTR} and injecting knowledge about entity types using learnable virtual tokens \cite{Chen2022KnowPrompt}. 
Our current work uses several relation templates designed in these studies.

\paragraph{LLMs for Relation Extraction.} In terms of exploring the RE capabilities of LLMs, most previous work has focused on investigating biomedical RE. 
\citet{Gutierrez2022GPT3BioIE} report that LLMs underperform standard small LMs fine-tuning in the few-shot setting on a comprehensive set of biomedical RE datasets and show evidence that the poor handling of the none-of-the-above (NoTA) relation category is one of the major culprits.
Furthermore, although a few RE-like tasks were included in Super Natural Instruction~\cite{Wang2022SuperInstructions}, these tasks constitute about $0.5$\% of the dataset and none of them were selected for model evaluation.


\section{Methodology}


In this section, we formally define the relation extraction problem and describe our multi-choice QA approach for the problem in detail.

\subsection{Problem Statement}
Relation extraction (RE) aims to extract the relationship between two given entities based on a specific sentence. 
More concretely, one relation example contains a sentence $S$ as well as a head entity $E_h$ and a tail entity $E_t$ within $S$. 
Given a relation example $(S, E_h, E_t)$, models are required to identify the relation between $E_h$ and $E_t$ expressed in the $S$ from a set of pre-defined relation types.

\subsection{Relation Templates}
Recent low-resource RE approaches~\cite{Sainz2021NLI, Lu2022SuRE, Han2022PTR} utilize relation-entailed templates as label verbalization (e.g., ``per:city\_of\_birth'' -> ``\{$E_h$\} was born in the city \{$E_t$\}''). 
As illustrated in Fig.~\ref{fig:intro} (left), the current SoTA method for low-resource RE ~\cite{Sainz2021NLI} utilizes manually constructed relation templates to reformulate the RE task as a natural language inference (NLI) task.

To ensure a fair comparison, we utilize the same templates developed in previous studies~\cite{Sainz2021NLI, Lu2022SuRE} to generate answer options within our QA4RE framework.
Furthermore, in Sec.~\ref{llm_nli} we discuss the possibility of directly applying the NLI formulation for RE in LLMs.

\subsection{QA4RE Framework}

As shown in Fig.~\ref{fig:intro} (right), we reformulate the relation extraction task as a multi-choice QA problem.
By integrating the given head and tail RE entities ($E_h$ and $E_t$) into the relation templates and using them as multiple-choice options, LLMs are able to leverage extensive QA instruction fine-tuning which has dramatically improved recent models. Additionally, our method allows LLM to generate only an answer index instead of the verbalized relation as in previous work~\cite{Gutierrez2022GPT3BioIE}, also shown in Fig.~\ref{fig:intro} (center).

\paragraph{Type-Constrained Answer Construction.}
To transform RE into a multiple-choice question, for a given relation example $(S, E_h, E_t)$, we utilize sentence $S$ as the context in standard QA and create options composed of pre-defined templates filled with $E_h$ and $E_t$ entities. 
To fairly compare with previous work, we apply type constraints (when applicable) to eliminate options for relation types that are not compatible with the entity types of the head and tail entities.
For instance, if the type of $E_h$ is PERSON, the relation ``org:country\_of\_headquarters'' would be deemed invalid given that a person does not have headquarters.




\section{Experiment Setup}

\subsection{Datasets}
We evaluate our methods on four RE datasets:
(1) TACRED~\cite{Zhang2017TACRED}, 
(2) RETACRED~\cite{Stoica2021RETACRED}, 
(3) TACREV~\cite{Alt2020TACREV}, and 
(4) SemEval 2010 Task 8 (SemEval for brevity)~\cite{Iris2010SemEval}.
Following previous work~\cite{Sainz2021NLI, Lu2022SuRE, Han2022PTR, Chen2022KnowPrompt}, we report the micro averaged F1 with the none-of-the-above relation excluded.
To keep OpenAI API costs under control, we randomly sample 1,000 examples from each dataset's test split as our test set.

\subsection{Baselines}
\paragraph{Zero-Shot.}
For small LM-based models, we evaluate two low-resource SoTA RE baselines: 
(1) As shown in Fig.~\ref{fig:intro} (left), NLI~\cite{Sainz2021NLI} reformulates RE as a natural language inference task and leverages several LMs fine-tuned on the MNLI dataset~\cite{Adina2018MNLI}: BART-Large~\cite{Lewis2020BART}, RoBERTa-Large~\cite{Liu2019RoBERTa}, and DeBERTa-XLarge~\cite{He2021DeBERTa}. 
This method holds the SoTA performance on both zero and few-shot RE.
(2) Besides, SuRE~\cite{Lu2022SuRE} frames RE as a summarization task and utilizes generative LMs such as BART-Large~\cite{Lewis2020BART} and PEGASUS-Large~\cite{Zhang2020Pegasus}, achieving competitive results in few-shot and fully-supervised settings.

For the NLI approach~\cite{Sainz2021NLI}, we report performance using their own templates on TACRED and TACREV. As this method does not have templates for RETACRED and SemEval, we use the templates from the follow-up work, SuRE~\cite{Lu2022SuRE}, on these two datasets instead.
All the zero-shot methods, including those on LLMs, apply entity type constraints to reduce the relation label space. 
Since SemEval does not provide entity types, the above methods use all possible relations in every instance as the label space.

\paragraph{Few-Shot.}
Though our main experiments focus on zero-shot RE, we further explore our method's capabilities by comparing their few-shot performance against several competitive small LM-based methods on the TACRED dataset.

The NLI baseline can be easily extended to the few-shot setting.\footnote{SuRE can also be extended to the few-shot setting but we were unable to replicate their results with the code provided.}
Furthermore, we add (1) standard Fine-Tuning~\cite{Gutierrez2022GPT3BioIE}, (2) PTR~\cite{Han2022PTR} using prompt-tuning with logical rules, and (3) KnowPrompt~\cite{Chen2022KnowPrompt} using entity type knowledge via learning virtual tokens, all of which are initialized with RoBERTa-Large~\cite{Liu2019RoBERTa}. For hyperparameter details, please refer to Appendix~\ref{sec:hp-for-fewshot-methods}.


\begin{table*}[]
\centering
\resizebox{\textwidth}{!}{
\begin{tabular}{ll|lllllllllllll}
\toprule
\multicolumn{2}{c|}{\multirow{2}{*}{\textbf{Methods}}}                  & \multicolumn{3}{l}{$\mspace{24mu}$ \textbf{TACRED}}                                                         & \multicolumn{3}{l}{$\mspace{13mu}$ \textbf{RETACRED}}                                                   & \multicolumn{3}{l}{$\mspace{24mu}$ \textbf{TACREV}}                                                        & \multicolumn{3}{l}{$\mspace{24mu}$ \textbf{SemEval}}                                                        & \multicolumn{1}{l}{$\mspace{1mu}$\textbf{Avg.}}            \\
\multicolumn{2}{c|}{}                                           & \multicolumn{1}{c}{P} & \multicolumn{1}{c}{R} & \multicolumn{1}{l}{$\mspace{3mu}$ F1}                      & \multicolumn{1}{c}{P} & \multicolumn{1}{c}{R} & \multicolumn{1}{l}{$\mspace{3mu}$ F1}                   & \multicolumn{1}{c}{P} & \multicolumn{1}{c}{R} & \multicolumn{1}{l}{$\mspace{3mu}$ F1}                     & \multicolumn{1}{c}{P} & \multicolumn{1}{c}{R} & \multicolumn{1}{l}{$\mspace{3mu}$ F1}                      & \multicolumn{1}{l}{$\mspace{3mu}$ F1}             \\ \midrule

\multicolumn{15}{l}{\textit{\textbf{Baselines}}}               \\ \midrule
\multicolumn{2}{l|}{NLI$_{\text{BART}}$}    & 42.6                  & 65.0                  & 51.4                                        & 59.5                  & 34.9                  & 44.0                                     & 44.0                  & 74.6                  & 55.3                                       & 21.6                  & 23.7                  & 22.6                                        & 43.3                                        \\
\multicolumn{2}{l|}{NLI$_{\text{RoBERTa}}$} & 37.1                  & 76.9                  & 50.1                                        & 52.3                  & 67.0                  & 58.7                                     & 37.1                  & 83.6                  & 51.4                                       & 17.6                  & 20.9                  & 19.1                                        & 44.8                                        \\
\multicolumn{2}{l|}{NLI$_{\text{DeBERTa}}$} & 42.9                  & 76.9                  & {\ul 55.1}                                  & 71.7                  & 58.3                  & 64.3                            & 43.3                  & 84.6                  & 57.2                                 & 22.0                  & 25.7                  & 23.7                                        & 50.1                                  \\

\multicolumn{2}{l|}{SuRE$_\text{BART}$}       & 13.1                  & 45.7                  & 20.4                                        & 17.9                  & 34.6                  & 23.6                                     & 14.1                  & 52.3                  & 22.2                                       & 0.0                   & 0.0                   & 0.0                                         & 16.5                                        \\
\multicolumn{2}{l|}{SuRE$_\text{PEGASUS}$}    & 13.8                  & 51.7                  & 21.8                                        & 16.6                  & 34.6                  & 22.4                                     & 13.5                  & 54.1                  & 21.6                                       & 0.0                   & 0.0                   & 0.0                                         & 16.4                                        \\ \midrule
\multicolumn{15}{l}{\textit{\textbf{GPT-3.5 Series}}}               \\ \midrule
\multirow{2}{*}{ChatGPT}             & Vanilla & 32.1 & 74.8 & 44.9                               & 45.4 & 61.3 & 52.1                                & 30.3 & 79.6 & 43.9                              & 18.2 & 20.8 & 19.4                                & 40.1                              \\
                                                & QA4RE   & 32.8 & 68.0 & 44.2 (\textcolor{WildStrawberry}{$-$0.7})    & 48.3 & 76.8 & 59.3 (\textcolor{ForestGreen}{$+$7.2})   & 34.7 & 79.1 & 48.2 (\textcolor{ForestGreen}{$+$4.3}) & 29.9 & 35.2 & 32.3  (\textcolor{ForestGreen}{$+$12.9}) & 46.0 (\textcolor{ForestGreen}{$+$5.9}) \\ \cdashlinelr{1-15}
\multirow{2}{*}{code-002}            & Vanilla          & 27.2                  & 70.1                  & 39.2                                        & 42.7                  & 70.4                  & 53.1                                     & 27.5                  & 77.7                  & 40.6                                       & 27.2                  & 25.6                  & 26.4                                        & 39.8                                        \\
                                             & QA4RE        & 37.7                  & 65.4                  & 47.8 (\textcolor{ForestGreen}{$+$8.6})           & 48.0                  & 74.0                  & 58.2  (\textcolor{ForestGreen}{$+$5.1})       & 31.7                  & 65.5                  & 42.7  (\textcolor{ForestGreen}{$+$2.1})         & 25.2                  & 29.2                  & 27.0  (\textcolor{ForestGreen}{$+$0.6})          & 43.9  (\textcolor{ForestGreen}{$+$4.1})          \\
\multirow{2}{*}{text-002}            & Vanilla          & 31.2                  & 73.1                  & 43.7                                        & 44.1                  & 76.3                  & 55.9                                     & 30.2                  & 76.8                  & 43.3                                       & 31.4                  & 28.8                  & 30.1                                        & 43.2                                        \\
                                             & QA4RE        & 35.6                  & 68.4                  & 46.8 (\textcolor{ForestGreen}{$+$3.1})           & 46.4                  & 72.4                  & 56.5 (\textcolor{ForestGreen}{$+$0.6})        & 35.7                  & 76.8                  & 48.8 (\textcolor{ForestGreen}{$+$5.4})          & 29.4                  & 34.3                  & 31.6  (\textcolor{ForestGreen}{$+$1.5})          & 45.9  (\textcolor{ForestGreen}{$+$2.7})          \\
\multirow{2}{*}{text-003}            & Vanilla          & 36.9                  & 68.8                  & 48.1                                        & 49.7                  & 62.2                  & 55.3                                     & 38.2                  & 76.8                  & 51.0                                 & 33.2                  & 39.3                  & 36.0                                  & 47.6                               \\
                                             & QA4RE        & 47.7                  & 78.6                  & \textbf{59.4} (\textcolor{ForestGreen}{$+$11.3}) & 56.2                  & 67.2                  & 61.2  (\textcolor{ForestGreen}{$+$5.9}) & 46.0                  & 83.6                  & \textbf{59.4} (\textcolor{ForestGreen}{$+$8.4}) & 41.7                  & 45.0                  & {\ul 43.3}  (\textcolor{ForestGreen}{$+$7.3}) & \textbf{55.8}  (\textcolor{ForestGreen}{$+$8.2}) \\ \midrule
\multicolumn{15}{l}{\textit{\textbf{FLAN-T5 Series}}}               \\ \midrule
\multirow{2}{*}{XLarge}  & Vanilla & 51.6 & 49.1 & 50.3                               & 54.3 & 40.3 & 46.3                                & 56.0 & 59.1 & {\ul 57.5}                               & 35.6 & 29.8 & 32.4                               & 46.6                              \\
                             & QA4RE   & 40.0 & 78.2 & 53.0  (\textcolor{ForestGreen}{$+$2.7}) & 57.1 & 79.7 & {\ul 66.5}  (\textcolor{ForestGreen}{$+$20.2}) & 40.7 & 85.9 & 55.3  (\textcolor{WildStrawberry}{$-$2.2})   & 45.1 & 40.1 & 42.5 (\textcolor{ForestGreen}{$+$10.1}) & 54.3 (\textcolor{ForestGreen}{$+$7.7}) \\
\multirow{2}{*}{XXLarge} & Vanilla & 52.1 & 47.9 & 49.9                               & 56.6 & 54.0 & 55.2                                & 52.6 & 50.9 & 51.7                               & 29.6 & 28.8 & 29.2                               & 46.5                              \\
                             & QA4RE   & 40.6 & 82.9 & 54.5 (\textcolor{ForestGreen}{$+$4.6})  & 56.6 & 82.9 & \textbf{67.3}  (\textcolor{ForestGreen}{$+$12.1}) & 39.6 & 86.4 & 54.3  (\textcolor{ForestGreen}{$+$2.6}) & 41.0 & 47.8 & \textbf{44.1} (\textcolor{ForestGreen}{$+$14.9}) & {\ul 55.1} (\textcolor{ForestGreen}{$+$8.6}) \\ \bottomrule
\end{tabular}
}
\caption{Experimental results on four RE datasets (\%). 
We omit the `davinci' within the names of GPT-3.5 Series LLMs and ChatGPT refers to gpt-3.5-turbo-0301.
We mark the best results in \textbf{bold}, the second-best {\ul underlined}, and F1 improvement of our QA4RE over vanilla RE in \textcolor{ForestGreen}{green}.}
\vspace{-5pt}
\label{tab:main-table}

\end{table*}

\subsection{QA4RE Implementation Details}
Our QA4RE framework utilizes the same templates and type constraints developed by prior work~\cite{Sainz2021NLI, Lu2022SuRE}.
In particular, we use SuRE~\cite{Lu2022SuRE} templates for our QA4RE approach on all 4 datasets since NLI~\cite{Sainz2021NLI} templates were only designed for TACRED.
For prompt engineering, we explore prompt formats and task instructions for vanilla RE and QA4RE in pilot experiments, using text-davinci-002 on a $250$-example subset of the TACRED dev set.
We then use the same task instructions and prompt format for all four datasets and LLMs.
Please refer to Appendix \ref{sec:prompts-for-llm} and \ref{sec:template} for prompt format and relation verbalization template details, respectively.

To systematically compare our QA4RE framework with the vanilla RE formulation, we evaluate them on two series of LLMs, resulting in seven models in total.
In GPT-3.5 series LLMs, for LLMs accessible via Text Completion API (code-davinci-002, text-davinci-002, and text-davinci-003), we follow previous work~\cite{Gutierrez2022GPT3BioIE} and use the logit bias option to constrain token generation to relation labels for vanilla RE and option indices for QA4RE. 
Due to the fewer available control options for LLMs in Chat Completion API (gpt-3.5-turbo-0301), we only set the temperature as 0 and use the default system prompt.

We also examine open-sourced FLAN-T5 series LLMs~\cite{ChungFlanT5} that are trained on a mixture of tasks~\cite{Sanh2022T0, Wei2022FLAN, Wang2022SuperInstructions}. 
The 1,836 tasks utilized in training include less than 0.5\% of RE-similar tasks, making FLAN-T5 series LLMs the ideal models for verifying our hypothesis.
Specifically, we use XLarge (3B) and XXLarge (11B) models and adopt the same prompts and greedy decoding strategy as GPT-3.5 series LLMs to ensure a fair comparison.



\section{Results}

\subsection{Zero-Shot Results}

Our main experimental results on four relation extraction datasets can be found in Tab.~\ref{tab:main-table}. 
We have the following observations from our results:

\noindent

\noindent
\textbf{(1)} 
By reformulating RE as QA, our framework improves upon the vanilla RE formulation on all the LLMs and most datasets, making them much stronger zero-shot relation extractors.
In particular, text-davinci-003 and FLAN-T5 XL and XXL are able to outperform the prior SoTA, NLI$_{\text{DeBERTa}}$, by a large margin.
One thing worth noting is that QA4RE brings the largest gain on the best LLM in each series  (text-davinci-003 and FLAN-T5 XXL), showing that stronger LLMs may benefit more from our framework.

\noindent
\textbf{(2)} 
The two FLAN-T5 LLMs in Tab.~\ref{tab:main-table} benefit significantly from our QA4RE framework. 
Moreover, consistent and substantial improvements can also be observed in other FLAN-T5 models and the full test set, as discussed in Sec.~\ref{sec:instruction-discussion} and Appendix~\ref{sec:full-test-flan-t5}. 
Considering that relation extraction tasks account for less than 0.5\% of the instruction tasks used to train FLAN-T5 models, these findings strongly support our hypothesis that aligning underrepresented tasks with more common instruction-tuning tasks, such as QA, unlocks LLMs' ability to solve low-frequency tasks.


\noindent
\textbf{(3)} The SemEval dataset poses a significant challenge for all baselines given its lack of type-constraints, particularly for SuRE~\cite{Lu2022SuRE}. 
With such a large search space, generative LMs without fine-tuning tend to summarize all examples into NoTA relation, resulting in its systematic failure.
It should be noted that without type constraints, the RE problem becomes a $19$-choice question answering task in our QA4RE framework. 
Despite this, our method still demonstrates substantial improvements for LLMs, particularly for text-davinci-003 and FLAN-T5 XXL.



\subsection{Robustness to Verbalization Templates}
For our experiments, we utilize manually written relation templates from previous work~\cite{Sainz2021NLI, Lu2022SuRE}. 
However, ~\citet{Lu2022SuRE} note that model performance may vary significantly with template design.
Thus, to investigate the robustness of models to different templates, thorough experiments are conducted with four different templates, described in detail in Appendix~\ref{sec:template}, across all zero-shot methods on the TACRED dataset.
Tab.~\ref{tab:template-robustness} shows results comparing these four templates on all methods used in our main experiments, including vanilla RE as a template-free reference.
\begin{table}[h]
\centering
\resizebox{0.47\textwidth}{!}{
\begin{tabular}{llcccc}
\toprule
\multicolumn{2}{c}{\textbf{Methods}}       & T\textsc{emp}1 & T\textsc{emp}2 & T\textsc{emp}3 & T\textsc{emp}4 \\ \midrule
\multicolumn{2}{l}{NLI$_{\text{BART}}$}    & 51.4           & 49.7           & 4.4            & 42.0           \\
\multicolumn{2}{l}{NLI$_{\text{RoBERTa}}$} & 50.1           & 47.1           & 19.6           & 35.8           \\
\multicolumn{2}{l}{NLI$_{\text{DeBERTa}}$} & 55.0           & 49.4           & 17.1           & 36.6           \\
\multicolumn{2}{l}{SuRE$_\text{BART}$}     & 19.9           & 20.4           & 2.1            & 10.1           \\
\multicolumn{2}{l}{SuRE$_\text{PEGASUS}$}  & 20.5           & 21.8           & 6.2            & 19.3           \\ \midrule
\multirow{2}{*}{text-003}      & Vanilla      & \multicolumn{4}{c}{48.1}                                          \\
                            & QA4RE   & \textbf{56.6}  & \textbf{59.4}  & \textbf{48.7}  & \textbf{50.1}  \\ \bottomrule
\end{tabular}
}
\caption{F1 score on TACRED with four templates (\%). The best result using each template is marked in bold. text-003 refers to text-davinci-003.}
\label{tab:template-robustness}
\end{table}

From Tab.~\ref{tab:template-robustness}, we observe the following:

\noindent
\textbf{(1)} Our method consistently outperforms small LM baselines and the vanilla RE framework, regardless of the template.
It is worth noting that even with templates that are constructed with label name information only and no expert knowledge (T\textsc{emp}3 and T\textsc{emp}4), our QA framework still performs better than vanilla RE, indicating the effectiveness and consistency of our QA framework.

\noindent
\textbf{(2)} NLI and SuRE performance is largely template dependent. When using carefully crafted high-quality templates (T\textsc{emp}1 and T\textsc{emp}2), several LM-based NLI methods outperform text-davinci-003 with vanilla RE. 
However, when equipped with templates created without expert knowledge (T\textsc{emp}3 and T\textsc{emp}4), the performance of both NLI and SuRE deteriorates dramatically.
In contrast, QA4RE is more robust to variation in verbalization templates, reducing trial-and-error development efforts as well as making it more readily transferred to settings where obtaining quality templates may be limited due to the high cost of expert annotations, such as the biomedical or financial domains.



\subsection{None-of-the-Above Relation Evaluation}
The none-of-the-above (NoTA) relation~\cite{Gao2019FewRel2,Sabo2021FS-TACRED, Gutierrez2022GPT3BioIE} is defined as the case where no relation of interest exists between the given entities. 
~\citet{Gutierrez2022GPT3BioIE} demonstrate that the earlier inferior performance of LLMs on RE tasks can be largely attributed to their inability to handle the NoTA relation.
To evaluate the efficacy of zero-shot methods on NoTA relation, following previous work~\cite{Fei2016BreakingClose, Shu2017DOC, Sainz2021NLI}, we apply NoTA-included macro F1 metric as well as micro and macro P vs.\ N (all positive relations vs. NoTA relation as binary classification) F1 metrics.

\begin{table}[th]
\centering
\small

\resizebox{\linewidth}{!}{
\begin{tabular}{llccc}
\toprule
\multicolumn{2}{c}{\textbf{Methods}}       & Macro F1 & Micro P vs.\ N & Macro P vs.\ N \\ \midrule
\multicolumn{2}{l}{NLI$_{\text{BART}}$}    & 49.8              & 75.9                  & 71.1                  \\
\multicolumn{2}{l}{NLI$_{\text{RoBERTa}}$} & 43.7              & 68.5                  & 65.8                  \\
\multicolumn{2}{l}{NLI$_{\text{DeBERTa}}$} & 55.0              & 75.6                  & 72.3                  \\
\multicolumn{2}{l}{SuRE$_\text{BART}$}     & 15.5              & 35.2                  & 35.0                  \\
\multicolumn{2}{l}{SuRE$_\text{PEGASUS}$}  & 14.9              & 32.4                  & 31.5                  \\ \midrule
\multirow{2}{*}{text-003}      & Vanilla      & 45.3              & 72.8                  & 69.5                  \\
                            & QA4RE        & \textbf{58.9}     & \textbf{78.4}         & \textbf{74.8}         \\ \bottomrule
\end{tabular}
}

\caption{NoTA-included $42$-class macro F1 as well as macro and micro P vs. N (all positive relations vs. NoTA) F1 on TACRED (\%). The best result of each metric is bolded. text-003 refers to text-davinci-003. Ma and Mi are short for macro and micro, respectively.}
\label{tab:nota-evaluation}
\end{table}

From Tab.~\ref{tab:nota-evaluation}, we observe that, when enhanced by our QA framework, text-davinci-003 achieves significant improvement in NoTA-included metrics, outperforming the small LM-based NLI methods. 
This further demonstrates the effectiveness of our framework, even in handling the challenging NoTA relation.
It is worth noting that these superior results are achieved by simply adding an entity-filled NoTA relation template as an answer option for QA, without the additional thresholding requirements of previous methods~\cite{Sainz2021NLI, Lu2022SuRE}. 
This eliminates the need for additional hyperparameter searching, which can be tricky for low-resource settings.

\subsection{Few-Shot Results}

While zero-shot RE is our main focus, we also evaluate our method under the few-shot setting. Results are shown in Tab.~\ref{tab:few-shot-comparison}.
Due to budget limitations, we restrict our case study to the $4$-shot scenario (i.e., $4$ labeled examples per relation) with the best-performing LLM in the zero-shot setting (text-davinci-003). 
After determining the optimal number of in-context examples searched on the dev set, we randomly select the examples with the same entity type constraints from the given train set.


Interestingly, vanilla RE is unable to obtain any improvement from labeled examples, suggesting that it is also limited in the few-shot setting.
The limited performance shown by vanilla RE indicates that few-shot demonstrations might bias the model towards incorrect relations in the context rather than helping it perform the task more accurately.

\begin{table}[!h]
\centering

\resizebox{0.47\textwidth}{!}{
\begin{tabular}{lccccc}
\toprule
\multicolumn{1}{c}{\textbf{Methods}}  & K=0  & K=4          & K=8          & K=16          & K=32 \\ \midrule
Fine-Tuning                           & -    & 9.0          & 21.2         & 29.3          & 33.9          \\
PTR                                   & -    & 26.8         & 30.0         & 32.9          & 36.8          \\
KnowPrompt                            & -    & 30.2         & 33.7         & 34.9          & 35.0          \\
NLI$_{\text{DeBERTa}}$-T\textsc{emp}1 & 55.0 & \textbf{64.2}& \textbf{64.7}& \textbf{58.7} & \textbf{65.7}          \\
NLI$_{\text{DeBERTa}}$-T\textsc{emp}2 & 49.4 & \textbf{51.2}&  47.3        & \textbf{50.5} & 48.1          \\ \midrule

Vanilla                               & 48.1 & 46.2         & \multicolumn{3}{c}{-}                        \\
QA4RE                                 & 59.4 & \textbf{62.0}& \multicolumn{3}{c}{-}                        \\ \bottomrule

\end{tabular}
}
\caption{Few-shot F1 on TACRED (\%). 
All results are averaged over $3$ different training subsets for each K. We use text-davinci-003 for vanilla RE and QA4RE. For the best-performing baseline (NLI) as well as vanilla RE and QA4RE, we mark the results in \textbf{bold} when they are improved over their zero-shot alternatives.}
\label{tab:few-shot-comparison}
\end{table}

Even employing our QA4RE framework, the few-shot text-davinci-003 does not outperform the DeBERTa-based NLI method \cite{Sainz2021NLI} when using their own templates (T\textsc{emp}1).
However, fine-tuning the NLI model on RE data can be brittle even with careful hyperparameter tuning, as evidenced by the unstable gains seen as more data is added for both T\textsc{emp}1 and T\textsc{emp}2. 
Furthermore, we find that few-shot NLI results when using T\textsc{emp}2 drop substantially from T\textsc{emp}1, suggesting that this approach also lacks robustness to templates in the few-shot setting. 
Thus, considering that our QA approach enables LLMs to obtain few-shot improvements over zero-shot results using random in-context learning example selection, obtains only around 2\% lower performance than the best NLI model, and is robust to different template designs, our approach is competitive on few-shot RE and has the potential to achieve even stronger performance with more exploration. We leave further investigation on how to improve LLMs for few-shot RE to future work.

\section{Discussions}

\subsection{Are Relation Templates All LLMs Need?}

\begin{figure}[!t]
 \centering
 \includegraphics[width=\columnwidth]{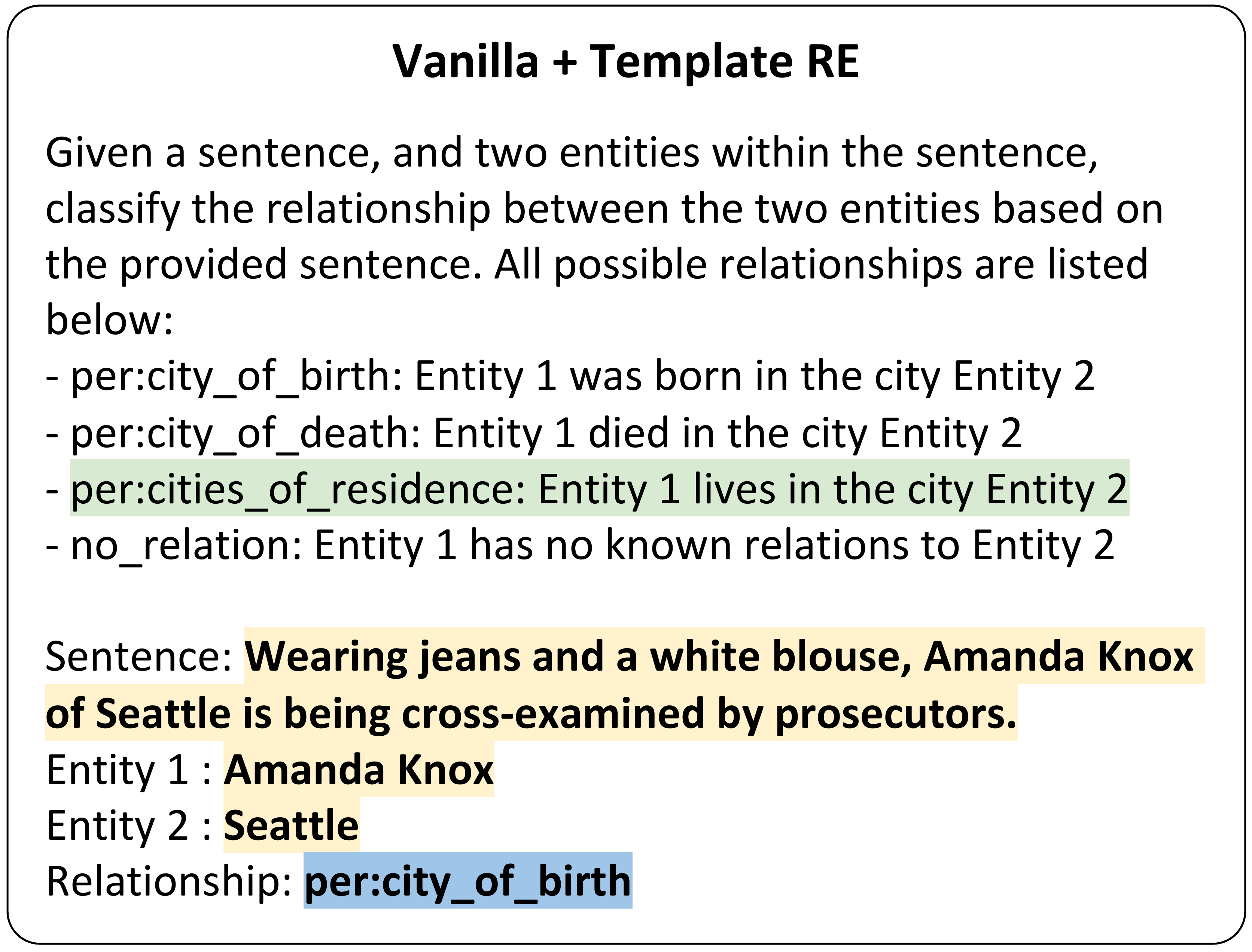}
\caption{The same example and templates as Fig.~\ref{fig:intro} but using templates for relation explanations.}
        \label{fig:explanation}
\end{figure}

We conduct an ablation study to better understand how relation templates contribute to the performance improvement obtained by QA4RE. As illustrated in Fig.~\ref{fig:explanation}, we fill the relation verbalization templates with markers \textit{Entity 1} and \textit{Entity 2} as relation explanations, thereby presenting the expert knowledge from the templates to the LLM.
Using the same templates and type constraints, we compare this framework (termed Vanilla+T\textsc{emp}) with vanilla RE and QA4RE on the TACRED dataset and GPT-3.5 series LLMs.

As shown in Tab.~\ref{tab:ablation}, introducing relation explanations using the same templates does not result in consistent or significant performance improvement. 
In fact, adding extra information to the task instruction might make it more challenging for the LLM to understand the task.
In contrast, using our QA4RE framework, we do not need to separately specify the entities of interest or relation explanations; they are both naturally embedded in the answer options.
These ablation results show that the gains from QA4RE mainly come from the QA reformulation, not simply from the relation explanations/templates.

\begin{table}[!t]
\centering
\resizebox{0.47\textwidth}{!}{
\begin{tabular}{llcccc}
\toprule
\multicolumn{2}{c}{\textbf{Methods}}            & P             & R             & F1            & $\Delta$F1   \\ \midrule
\multirow{3}{*}{code-002} & Vanilla     & 27.2          & 70.1          & 39.2          & -             \\
                                  & Vanilla + T\textsc{emp} & 27.5          & 71.8          & 39.7          & \textcolor{ForestGreen}{$+$0.5}           \\
                                  & QA4RE  & 37.7          & 65.4          & 47.8          & \textcolor{ForestGreen}{$+$8.6}           \\ \midrule
\multirow{3}{*}{text-002} & Vanilla     & 31.2          & 73.1          & 43.7          & -             \\
                                  & Vanilla + T\textsc{emp} & 26.8          & 77.8          & 39.8          & \textcolor{WildStrawberry}{$-$3.9}          \\
                                  & QA4RE  & 35.6          & 68.4          & 46.8          & \textcolor{ForestGreen}{$+$3.1}           \\ \midrule
\multirow{3}{*}{text-003} & Vanilla     & 36.9          & 68.8          & 48.1          & -             \\
                                  & Vanilla + T\textsc{emp} & 36.9          & 76.5          & 49.8          & \textcolor{ForestGreen}{$+$1.7}           \\
                                  & QA4RE  & \textbf{47.7} & \textbf{78.6} & \textbf{59.4} & \textcolor{ForestGreen}{$+$11.3} \\ \bottomrule
\end{tabular}
}
\caption{Evaluation on TACRED regarding whether incorporating relation explanations based on the same templates into vanilla RE bridges its gap to QA4RE (\%).}
\label{tab:ablation}
\vspace{-10pt}
\end{table}


\subsection{QA4RE vs. NLI4RE}\label{llm_nli}

Given the strong performance obtained by small LMs using the NLI reformulation of RE, we leverage this same formulation~\cite{Sainz2021NLI} for LLMs (termed NLI4RE).\footnote{We follow the NLI format from ANLI \cite{Wang2022SuperInstructions}.}
More concretely, for each example, we use the LLM to predict whether the given sentence (the premise) entails each answer option from the QA4RE formulation (the hypothesis). We allow the LLM to generate \textit{entailment}, \textit{neutral}, or \textit{contradiction} for each sentence-relation pair. 
If the maximum probability of entailment among all possible positive relations is below the threshold of 0.5, the example will be classified as NoTA, as done by \citet{Sainz2021NLI}.

\begin{table}[!h]
\small
\centering
\resizebox{0.47\textwidth}{!}{
\begin{tabular}{l|ccccc}
\toprule
\textbf{Formulation} & \multicolumn{1}{l}{\textbf{RED}} & \textbf{RERED} & \textbf{REV} & \textbf{Eval} & \textbf{Avg.} \\ \midrule
Vanilla      & 48.1                                & 55.3               & 51.0            & 36.0             & 47.6          \\
NLI4RE       & 41.7                                & 36.8               & 39.2            & 22.4             & 35.0          \\
QA4RE        & \textbf{59.4}                       & \textbf{61.2}      & \textbf{59.4}   & \textbf{43.3}    & \textbf{55.8} \\ \bottomrule
\end{tabular}
}
\caption{F1 of text-davinci-003 with different task formulations (\%). RED, RERED, REV, and Eval are short for TACRED, RETACRED, TACREV, and SemEval datasets, respectively.}
\label{tab:formulation-comparison}
\vspace{-5pt}
\end{table}

As shown in Tab.~\ref{tab:formulation-comparison}, when using the NLI formulation, text-davinci-003 surprisingly underperforms the vanilla RE formulation. 
The reason for its poor performance is two-fold: (1) The heuristically pre-defined threshold $0.5$ is not ideal for LLMs and thus many positive predictions are classified as NoTA.
However, it is also difficult to find a good threshold under the zero-shot setting.
(2) Under NLI4RE, unlike vanilla RE or QA4RE, an LLM is not seeing the full relation space but assigning probabilities to each candidate hypothesis individually.
The final prediction is thus more sensitive to the LLM's bias over different relations.

NLI4RE also requires multiple inference runs for each relation example to evaluate all the candidate relations, incurring a significantly higher cost.

\subsection{QA4RE \& Model Size}

\begin{table}[!t]
\small
\centering
\resizebox{0.47\textwidth}{!}{
\begin{tabular}{lcccc}
\toprule
\multicolumn{1}{c}{\multirow{2}{*}{\textbf{LMs}}} & \multicolumn{1}{c}{\multirow{2}{*}{\textbf{Model Size}}} & \multicolumn{3}{c}{\textbf{Avg. F1}}                                                                                               \\
\multicolumn{1}{c}{}                                       & \multicolumn{1}{c}{}                                   & Vanilla          & QA4RE          & $\Delta$ \\  \midrule
\textit{\textbf{GPT-3.5 Series}} & & & &  \\\midrule
text-001                                                       & 175B                                                   & 22.3          & 14.9       & \textcolor{WildStrawberry}{$-$7.4} \\
code-002                                                      & 175B                                                   & 39.8          & 43.9       & \textcolor{ForestGreen}{$+$4.1} \\
text-002                                                      & 175B                                                     & 43.2          & 45.9       & \textcolor{ForestGreen}{$+$2.7} \\
text-003                                                      & 175B                                                    & 47.6          & 55.8       & \textcolor{ForestGreen}{$+$8.2} \\ \midrule
\textit{\textbf{FLAN-T5 Series}} & & & &  \\\midrule
Small                                                      & 80M & 19.5             & 25.0           & \textcolor{ForestGreen}{$+$5.6}                            \\
Base                                                       & 250M & 22.3             & 26.4           & \textcolor{ForestGreen}{$+$4.2}                            \\
Large                                                      & 780M & 34.8             & 41.8           & \textcolor{ForestGreen}{$+$7.0}                            \\
XLarge                                                     & 3B & 46.6             & 54.3           & \textcolor{ForestGreen}{$+$7.7}                        \\
XXLarge                                                    & 11B & 46.5             & 55.1           & \textcolor{ForestGreen}{$+$8.6}                            \\
\bottomrule
\end{tabular}
}
\caption{
Effectiveness of QA4RE on both the GPT-3.5 series and FLAN-T5 with different sizes. The results are averaged over four RE datasets.
}
\label{tab:llm-sizes}
\vspace{-10pt}
\end{table}
\label{sec:instruction-discussion}

\noindent
To verify the effectiveness and transferability of our QA4RE framework on smaller instruction-tuned models, we further evaluate the FLAN-T5 Small (80M), Base (250M), and Large (780M) on the full test set over four RE datasets.
Tab.~\ref{tab:llm-sizes} shows our QA4RE framework can still bring considerable gains to instruction-tuned models with various sizes, even for the smallest one (80M).
This demonstrates the effectiveness of QA4RE is transferable across various model sizes from 80M to 175B, considering the consistent improvements of QA4RE on several GPT-3.5 models.

In the FLAN-T5 series, larger models benefit more from our framework.
However, we note that this trend does not continue when scaling up to much larger GPT-3.5 models.
In fact, all GPT-3.5 models except for text-davinci-003 benefit less from QA4RE than FLAN-T5 models.
The smaller improvements of QA4RE on these models make their overall RE performance only comparable with models that are approximately 20 and 50 times smaller.
This indicates that the wide variety of alignment strategies used by the GPT-3.5 series models discussed in Sec.~\ref{sec:related_work} might not be universally more effective than standard instruction-tuning for improving model generalization on low-incidence tasks even when aligned to high incidence ones.
Nevertheless, the strong improvement observed in the strongest models tested, text-davinci-003 and FLAN-T5-XXL, demonstrates the potential for QA4RE's effectiveness to continue as models become even more capable in the future.

\section{Conclusions and Future Work}

In this work, we first show that even the most recent instruction-tuned LLMs underperform fine-tuned small LMs on the relation extraction (RE) task.
To address this limitation, we reformulate RE into multiple-choice question answering (QA) with the purpose of leveraging a task that is widely covered in instruction-tuning datasets like QA, instead of RE, which is barely present in these datasets. 
Comprehensive experiments demonstrate that our QA4RE framework unlocks the power of LLMs as zero-shot relation extractors, especially for two recent LLMs (text-davinci-003 and FLAN-T5 XXL).
We also conduct thorough experiments to explore the robustness and few-shot effectiveness of our method as well as study in what LLM training scenarios it is most effective.

In future work, we hope to explore additional underrepresented tasks in instruction-tuning that might be challenging for LLMs and could be successfully aligned with more widely adopted instruction-tuning tasks like QA. 
Additionally, we plan to continue exploring this line of work by leveraging our QA4RE framework for other LLMs such as the OPT-series~\cite{Zhang2022OPT, Iyer2022OPT-IML} and PaLM~\cite{Chowdhery2022PaLM}, which are not included in this work due to the limited computational resources and/or access.



%


\section{Limitations}
%

Even though our method helps unleash the power of six recent strong LLMs as zero-shot relation extractors, earlier LLMs without strong instruction tuning such as text-davinci-001 saw no improvements from our framework.
Additionally, although we carry out comprehensive experiments on the zero-shot RE setting, our few-shot exploration is more limited.
It is still unclear from our investigation whether including even more training examples can improve LLM's RE performance and to what extent the same trends seen across GPT-3 models in the zero-shot setting hold steady in the few-shot setting. 
We leave answering these questions for future work.

\section{Ethics Statement}
In this work, we propose a method to improve LLM performance on the important and fundamental task of relation extraction.
We do not anticipate any ethical issues regarding the topics of this research.

\section*{Acknowledgements}
The authors would like to thank Renze Lou, colleagues from the
OSU NLP group, and the anonymous reviewers for their valuable feedback. 
The authors would also like to thank Keming Lu for discussions and guidance on reproducing SuRE.
This research was supported in part by NSF OAC 2112606, NIH R01LM014199, and Ohio Supercomputer Center \citep{OhioSupercomputerCenter1987}.

\bibliography{custom}
\bibliographystyle{acl_natbib}


\appendix
\label{sec:appendix}
\clearpage

\begin{table*}[t]
\centering
\resizebox{\textwidth}{!}{
\begin{tabular}{ll|lllllllllllll}
\toprule
\multicolumn{2}{c|}{\multirow{2}{*}{\textbf{Methods}}}                  & \multicolumn{3}{l}{$\mspace{24mu}$ \textbf{TACRED}}                                                         & \multicolumn{3}{l}{$\mspace{13mu}$ \textbf{RETACRED}}                                                   & \multicolumn{3}{l}{$\mspace{24mu}$ \textbf{TACREV}}                                                        & \multicolumn{3}{l}{$\mspace{24mu}$ \textbf{SemEval}}                                                        & \multicolumn{1}{l}{$\mspace{1mu}$\textbf{Avg.}}            \\
\multicolumn{2}{c|}{}                                           & \multicolumn{1}{c}{P} & \multicolumn{1}{c}{R} & \multicolumn{1}{l}{$\mspace{3mu}$ F1}                      & \multicolumn{1}{c}{P} & \multicolumn{1}{c}{R} & \multicolumn{1}{l}{$\mspace{3mu}$ F1}                   & \multicolumn{1}{c}{P} & \multicolumn{1}{c}{R} & \multicolumn{1}{l}{$\mspace{3mu}$ F1}                     & \multicolumn{1}{c}{P} & \multicolumn{1}{c}{R} & \multicolumn{1}{l}{$\mspace{3mu}$ F1}                      & \multicolumn{1}{l}{$\mspace{3mu}$ F1}             \\ \midrule

\multirow{2}{*}{Small}        & Vanilla      & 9.5                   & 40.9                  & 15.4                                    & 22.8                  & 50.2                  & 31.3                                    & 9.1                   & 41.9                  & 15.0                                      & 10.0                  & 11.8                  & 10.8                                    & 18.1                                   \\
                              & QA4RE        & 13.8                  & 52.2                  & 21.8 (\textcolor{ForestGreen}{$+${6.4}})  & 33.5                  & 66.2                  & 44.5 (\textcolor{ForestGreen}{$+${13.2}}) & 13.7                  & 55.2                  & 22.0 (\textcolor{ForestGreen}{$+${7.0}})    & 5.9                   & 7.1                   & 6.4 (\textcolor{WildStrawberry}{$-${4.4}})   & 23.7 (\textcolor{ForestGreen}{$+${5.6}}) \\
\multirow{2}{*}{Base}         & Vanilla      & 14.1                  & 31.1                  & 19.4                                    & 21.1                  & 26.8                  & 23.6                                    & 14.1                  & 33.3                  & 19.8                                      & 14.9                  & 17.9                  & 16.2                                    & 19.8                                   \\
                              & QA4RE        & 17.1                  & 54.7                  & 26.0 (\textcolor{ForestGreen}{$+${6.6}})  & 33.0                  & 65.2                  & 43.8 (\textcolor{ForestGreen}{$+${20.2}}) & 17.2                  & 58.5                  & 26.6 (\textcolor{ForestGreen}{$+${6.8}})    & 6.7                   & 8.0                   & 7.3 (\textcolor{WildStrawberry}{$-${8.9}})   & 25.9 (\textcolor{ForestGreen}{$+${6.2}}) \\
\multirow{2}{*}{Large}        & Vanilla      & 22.8                  & 58.6                  & 32.8                                    & 37.5                  & 60.8                  & 46.4                                    & 22.6                  & 61.9                  & 33.1                                      & 23.7                  & 19.7                  & 21.5                                    & 33.5                                   \\
                              & QA4RE        & 30.3                  & 78.5                  & 43.7 (\textcolor{ForestGreen}{$+${10.9}}) & 44.5                  & 72.6                  & 55.2 (\textcolor{ForestGreen}{$+${8.8}})  & 29.9                  & 82.4                  & 43.9 (\textcolor{ForestGreen}{$+${10.8}})   & 24.8                  & 15.8                  & 19.3 (\textcolor{WildStrawberry}{$-${2.2}})  & 40.5 (\textcolor{ForestGreen}{$+${7.1}}) \\
\multirow{2}{*}{XLarge}       & Vanilla      & 48.8                  & 49.0                  & 48.9                                    & 55.8                  & 39.8                  & 46.4                                    & 52.0                  & 55.7                  & \textbf{53.8}                                      & 34.9                  & 29.6                  & 32.0                                    & 45.3                                   \\
                              & QA4RE        & 37.6                  & 78.6                  & {\ul 50.9} (\textcolor{ForestGreen}{$+${2.0}})  & 56.2                  & 79.9                  & {\ul 66.0} (\textcolor{ForestGreen}{$+${19.6}}) & 38.2                  & 84.7                  & 52.7 (\textcolor{WildStrawberry}{$-${1.1}}) & 44.4                  & 39.9                  & {\ul 42.1} (\textcolor{ForestGreen}{$+${10.1}}) & {\ul 52.9} (\textcolor{ForestGreen}{$+${7.7}}) \\
\multirow{2}{*}{XXLarge}      & Vanilla      & 48.2                  & 45.3                  & 46.7                                    & 56.1                  & 53.7                  & 54.9                                    & 50.6                  & 50.6                  & 50.6                                      & 29.2                  & 28.1                  & 28.6                                    & 45.2                                   \\
                              & QA4RE        & 38.1                  & 82.9                  & \textbf{52.2} (\textcolor{ForestGreen}{$+${5.5}})  & 55.9                  & 82.0                  & \textbf{66.5} (\textcolor{ForestGreen}{$+${11.6}}) & 38.3                  & 88.1                  & {\ul 53.4} (\textcolor{ForestGreen}{$+${2.8}})    & 40.2                  & 47.5                  & \textbf{43.5} (\textcolor{ForestGreen}{$+${14.9}}) & \textbf{53.9} (\textcolor{ForestGreen}{$+${8.7}}) \\ \bottomrule
\end{tabular}
}
\caption{FLAN-T5 results on full test set of four RE datasets (\%). 
We mark the best results in \textbf{bold}, the second-best {\ul underlined}, and F1 improvement of our QA4RE over vanilla RE in \textcolor{ForestGreen}{green}.}
\vspace{-5pt}
\label{tab:appendix-full-FLAN-T5-results}

\end{table*}

\section{Instruction Dataset Portion}
\label{sed:appendix-instruction-dataset-portion}

\begin{table}[H]
\centering
\resizebox{0.47\textwidth}{!}{
\begin{tabular}{lccc}
\toprule
                    & \#Tasks & \%RE & \%QA \\ \midrule
T0~\cite{Sanh2022T0} & 62      & 0    & 27.4 \\
FLAN~\cite{Wei2022FLAN}& 62      & 0    & 21   \\
MetaICL~\cite{Min2022MetaICL}& 142     & 0    & 28.9 \\
NaturalInstruct~\cite{Wang2022SuperInstructions} & 1731  & $<$0.5 & $>$12  \\ \bottomrule
\end{tabular}
}
\caption{Popular instruction tuning datasets and proportion of RE and QA tasks in each.}
\label{tab:appendix-dataset-portion}
\end{table}
As shown in Tab.~\ref{tab:appendix-dataset-portion}, there is no RE task in T0~\cite{Sanh2022T0}, FLAN~\cite{Wei2022FLAN}, and MetaICL~\cite{Min2022MetaICL} instruction tuning datasets. 
Even in the largest available NaturalInstruct~\cite{Wang2022SuperInstructions}, RE tasks consist of only less than 0.5\% of the total tasks. 
By contrast, QA is the most popular task format in all instruction tuning datasets. 
These observations indicate the low incidence of RE tasks and the dominance of QA tasks in datasets used for instruction tuning.

\section{Experimental Details}

\subsection{Hyperparameters for Few-Shot Methods}

In the few-shot setting, for each K, we randomly sample $3$ times to obtain different training subsets, each of which will be used as in-context demonstrations for LLMs or used to train the small language models in baselines. Report results are averaged over the three subsets. To avoid over-estimating few-shot performance with too many dev examples~\cite{Perez2021TrueFL}, we use $100$ randomly selected examples of dev set for all the hyperparameter searching.

For LLMs, we use the dev set to search for the optimal number of in-context examples as a hyperparameter from $\{1, 2, 5\}$.
Then we randomly select the same type-constrained in-context examples from the given train set.

For all small LM-based baselines, we use their publicly available code and hyper-parameters for training.
\label{sec:hp-for-fewshot-methods}
According to the original papers of NLI~\cite{Sainz2021NLI} and SuRE~\cite{Lu2022SuRE}, we use the checkpoints available online and hyperparameters reported for model training.
Unfortunately, we were unable to reproduce SuRE results with default hyperparameters.
For standard Fine-Tuning~\cite{Gutierrez2022GPT3BioIE}, PTR~\cite{Han2022PTR}, and KnowPrompt~\cite{Chen2022KnowPrompt},
we perform a grid search over hyperparameters on dev with the range shown in Tab.~\ref{tab:appendix-few-shot-search-space}.

We use $8$ NVIDIA GeForce RTX 2080 Ti and $2$ NVIDIA RTX A6000 to conduct all the experiments. The total GPU hours used and the cost for OpenAI API are listed in Tab.~\ref{tab:appendix-total-cost}.

\begin{table}[H]
\centering
\small
\begin{tabular}{lc}
\toprule
\textbf{Hyperparameter}                                                                   & \textbf{Search Space}                               \\ \midrule
Learning Rate 1:                                                                          & \{$1$e$-5$, $3$e$-5$\}                                      \\
Weight Decay:                                                                             & \{$0.01$, $0.001$\}                                     \\
Learning Rate 2:                                                                          & \{$5$e$-5$, $2$e$-4$\}                                      \\
\bottomrule
\end{tabular}
\caption{Hyperparameters used for grid search of few-shot methods. Learning Rate 2 is used for training new tokens in PTR~\cite{Han2022PTR} and virtual tokens in KnowPrompt~\cite{Chen2022KnowPrompt}.}
\label{tab:appendix-few-shot-search-space}
\end{table}
\begin{table}[H]
\centering
\resizebox{0.47\textwidth}{!}{
\begin{tabular}{lccc}
\toprule
                        & \textbf{Num of Params}   & \multicolumn{1}{c}{\textbf{Total GPU}} & \multicolumn{1}{c}{\textbf{Total}} \\
\textbf{}               & \textbf{(Millions)}       & \multicolumn{1}{c}{\textbf{Hours}}     & \multicolumn{1}{c}{\textbf{Cost}}  \\ \midrule
\textbf{RoBERTa-Large}  & $354$                    &  $284$                                 & -                                  \\
\textbf{DeBERTa-XLarge} & $900$                    &  $14$                                  & -                                  \\
\textbf{BART-Large}     & $406$                    &  $2$                                   & -                                  \\
\textbf{Pegasus-Large}  & $568$                    &  $50$                                  & -                                  \\
\midrule
\textbf{FLAN-T5 S}      & $80$                     &  <1                                  & -                                  \\
\textbf{FLAN-T5 M}      & $250$                    &  <1                                  & -                                  \\
\textbf{FLAN-T5 L}      & $780$                    &  1                                  & -                                  \\
\textbf{FLAN-T5 XL}     & $3,000$                  &  2                                  & -                                  \\
\textbf{FLAN-T5 XXL}    & $11,000$                 &  4                                  & -                                  \\
\midrule
\textbf{OpenAI Text API}   & $175,000$                  & -                                      & ~\$$835$                              \\
\textbf{OpenAI Chat API}   & $?$                  & -                                      & ~\$$4$                              \\
\bottomrule
\end{tabular}
}
\caption{Total GPU Hours for open sources LMs and cost for using OpenAI API (all version included).}
\label{tab:appendix-total-cost}
\end{table}

\subsection{Prompts for LLMs}

\label{sec:prompts-for-llm}

As shown in Tab.~\ref{tab:appendix-prompt-design}, we list all templates used in this paper including vanilla + T\textsc{emp} in Tab.~\ref{tab:ablation}, NLI4RE in Tab.~\ref{tab:formulation-comparison}, and vanilla as well as QA4RE in all experiments.

\subsection{Relation Verbalization Templates}
\label{sec:template}
In the relation verbalization template robustness experiment shown in Tab.~\ref{tab:template-robustness}, the differences between four templates are described below using the \textit{org:top\_members/employees} relation from TACRED benchmark as an example:
\begin{enumerate}
    \item Concrete Examples: \textit{\{$E_h$\} is a chairman/\\president/director of \{$E_t$\}}
    \item Semantic Relationship: \textit{\{$E_h$\} is a high level member of \{$E_t$\}}
    \item Straightforward: \textit{The relation between \{$E_h$\} and \{$E_t$\} is top members or employees}
    \item Word Translation: \textit{\{$E_h$\} organization top members or employees \{$E_t$\}}
\end{enumerate}
The first set of templates was written by \citet{Sainz2021NLI}, while the remaining three were explored by \citet{Lu2022SuRE}. 
We use the templates from their official GitHub repositories.\footnote{Templates for Robustness Experiments:\\
T\textsc{emp}1: \href{
https://github.com/osainz59/Ask2Transformers/blob/master/resources/predefined\_configs/tacred.relation.config.json
}{\text{https://github.com/osainz59/Ask2Transformers/blob/}
master/resources/predefined\_configs/tacred.relation.config.json}\\
T\textsc{emp}3: \href{https://github.com/luka-group/SuRE/blob/main/data/templates/tacred/rel2temp\_forward.json}{\text{https://github.com/luka-group/SuRE/blob/main/data}\\
templates/tacred/rel2temp\_forward.json} \\
T\textsc{emp}4: \href{https://github.com/luka-group/SuRE/blob/main/data
/templates/tacred/rel2temp\_raw\_relation.json}{\text{https://github.com/luka-group/SuRE/blob/main/data}\\
/templates/tacred/rel2temp\_raw\_relation.json} \\
}
In addition, we further list relation verbalization templates used by all LLMs in our paper in Tab.~\ref{tab:appendix-sure-tacred-template}, Tab.~\ref{tab:appendix-sure-retacred-template}, and Tab.~\ref{tab:appendix-sure-semeval-template}.

\section{Full Test Results on FLAN-T5}
\label{sec:full-test-flan-t5}



We present the full test set results of all four RE datasets in Tab.~\ref{tab:appendix-full-FLAN-T5-results}. 
Our observations align with the findings from experiments on 1,000 test examples:

\noindent
\textbf{(1)} Our QA4RE framework can bring consistent and significant improvements over all FLAN-T5 series models on the averaged results. Additionally, larger models benefit more from our framework. 
These two signals strongly demonstrate the effectiveness of QA4RE.

\noindent
\textbf{(2)} We notice that our QA4RE does not improve smaller versions of FLAN-T5 on SemEval, a 19-choice QA task. 
This may be due that these models have difficulties in understanding long input fed by QA4RE.






\begin{table*}[]
\resizebox{\textwidth}{!}{
\centering
\small
\begin{tabular}{p{0.12\textwidth} | p{0.85\textwidth}}
\toprule
\textbf{Formulations} & \multicolumn{1}{c}{\textbf{Prompts}}                                                                                                                                                                                                                                                                                                                                                                                                                                                                                \\ \midrule
Vanilla RE            & \begin{tabular}[c]{p{0.85\textwidth}}Given a sentence, and two entities within the sentence, classify the relationship between the two entities based on the provided sentence. All possible Relationships are listed below:\\ - [Possible Relation 1]\\ - [Possible Relation 2]\\ - [NoTA Relation]\\ \\ Sentence: [Sentence $S$]\\ Entity 1: [Head Entity $E_{h}$]\\ Entity 2: [Tail Entity $E_{t}$]\\ Relationship:\end{tabular}                                                                         \\ \midrule
Vanilla + T\textsc{emp}& \begin{tabular}[c]{p{0.85\textwidth}}Given a sentence, and two entities within the sentence, classify the relationship between the two entities based on the provided sentence. All possible Relationships are listed below with explanations:\\ - [Possible Relation 1]: [Relation 1 Template]\\ - [Possible Relation 2]: [Relation 2 Template]\\ - [NoTA Relation]: [NoTA Relation Template]\\ \\ Sentence: [Sentence $S$]\\ Entity 1: [Head Entity $E_{h}$]\\ Entity 2: [Tail Entity $E_{t}$]\\ Relationship:\end{tabular} \\ \midrule
NLI4RE                & \begin{tabular}[c]{p{0.85\textwidth}}In this task, you will be presented with a premise and a hypothesis sentence. \\ Determine whether the hypothesis sentence entails (implies), contradicts (opposes), or is neutral with respect to the given premise sentence. Please answer with "Contradiction", "Neutral", or "Entailment".\\ \\ Premise: [Sentence $S$]\\ Hypothesis: [Entities in Relation 1 Template]\\ \\ Category:\end{tabular}                                                                            \\ \midrule
QA4RE                 & \begin{tabular}[c]{p{0.85\textwidth}}Determine which option can be inferred from the given Sentence.\\ \\ Sentence: [Sentence $S$]\\ Options:\\ A. [Entities in Relation 1 Template]\\ B. [Entities in Relation 2 Template]\\ C. [Entities in NoTA Relation Template]\\ \\ Which option can be inferred from the given Sentence?\\ Option:\end{tabular}   \\ \bottomrule                                                                                                                                                           
\end{tabular}
}
\caption{Prompt Formats of frameworks for LLMs in this paper. We only demonstrate NLI4RE with 1 template for simplicity.}
\label{tab:appendix-prompt-design}

\end{table*}
\begin{table*}[]
\centering
\small
\begin{tabular}{ll}
\toprule
\textbf{Relation}                   & \textbf{Template}                                          \\ \midrule
no\_relation                          & \{$E_h$\} has no known relations to \{$E_t$\}                  \\
per:stateorprovince\_of\_death        & \{$E_h$\} died in the state or province \{$E_t$\}              \\
per:title                             & \{$E_h$\} is a \{$E_t$\}                                       \\
org:member\_of                        & \{$E_h$\} is the member of \{$E_t$\}                           \\
per:other\_family                     & \{$E_h$\} is the other family member of \{$E_t$\}              \\
org:country\_of\_headquarters         & \{$E_h$\} has a headquarter in the country \{$E_t$\}           \\
org:parents                           & \{$E_h$\} has the parent company \{$E_t$\}                     \\
per:stateorprovince\_of\_birth        & \{$E_h$\} was born in the state or province \{$E_t$\}          \\
per:spouse                            & \{$E_h$\} is the spouse of \{$E_t$\}                           \\
per:origin                            & \{$E_h$\} has the nationality \{$E_t$\}                        \\
per:date\_of\_birth                   & \{$E_h$\} has birthday on \{$E_t$\}                            \\
per:schools\_attended                 & \{$E_h$\} studied in \{$E_t$\}                                 \\
org:members                           & \{$E_h$\} has the member \{$E_t$\}                             \\
org:founded                           & \{$E_h$\} was founded in \{$E_t$\}                             \\
per:stateorprovinces\_of\_residence   & \{$E_h$\} lives in the state or province \{$E_t$\}             \\
per:date\_of\_death                   & \{$E_h$\} died in the date \{$E_t$\}                           \\
org:shareholders                      & \{$E_h$\} has shares hold in \{$E_t$\}                         \\
org:website                           & \{$E_h$\} has the website \{$E_t$\}                            \\
org:subsidiaries                      & \{$E_h$\} owns \{$E_t$\}                                       \\
per:charges                           & \{$E_h$\} is convicted of \{$E_t$\}                            \\
org:dissolved                         & \{$E_h$\} dissolved in \{$E_t$\}                               \\
org:stateorprovince\_of\_headquarters & \{$E_h$\} has a headquarter in the state or province \{$E_t$\} \\
per:country\_of\_birth                & \{$E_h$\} was born in the country \{$E_t$\}                    \\
per:siblings                          & \{$E_h$\} is the siblings of \{$E_t$\}                         \\
org:top\_members/employees            & \{$E_h$\} has the high level member \{$E_t$\}                  \\
per:cause\_of\_death                  & \{$E_h$\} died because of \{$E_t$\}                            \\
per:alternate\_names                  & \{$E_h$\} has the alternate name \{$E_t$\}                     \\
org:number\_of\_employees/members     & \{$E_h$\} has the number of employees \{$E_t$\}                \\
per:cities\_of\_residence             & \{$E_h$\} lives in the city \{$E_t$\}                          \\
org:city\_of\_headquarters            & \{$E_h$\} has a headquarter in the city \{$E_t$\}              \\
per:children                          & \{$E_h$\} is the parent of \{$E_t$\}                           \\
per:employee\_of                      & \{$E_h$\} is the employee of \{$E_t$\}                         \\
org:political/religious\_affiliation  & \{$E_h$\} has political affiliation with \{$E_t$\}             \\
per:parents                           & \{$E_h$\} has the parent \{$E_t$\}                             \\
per:city\_of\_birth                   & \{$E_h$\} was born in the city \{$E_t$\}                       \\
per:age                               & \{$E_h$\} has the age \{$E_t$\}                                \\
per:countries\_of\_residence          & \{$E_h$\} lives in the country \{$E_t$\}                       \\
org:alternate\_names                  & \{$E_h$\} is also known as \{$E_t$\}                           \\
per:religion                          & \{$E_h$\} has the religion \{$E_t$\}                           \\
per:city\_of\_death                   & \{$E_h$\} died in the city \{$E_t$\}                           \\
per:country\_of\_death                & \{$E_h$\} died in the country \{$E_t$\}                        \\
org:founded\_by                       & \{$E_h$\} was founded by \{$E_t$\}                             \\ \bottomrule
\end{tabular}
\caption{Templates for TACRED and TACREV datasets.}
\label{tab:appendix-sure-tacred-template}
\end{table*}
\begin{table*}[]
\centering
\small
\begin{tabular}{ll}
\toprule
\textbf{Relation}                    & \textbf{Template}                                                                            \\ \midrule
no\_relation                         & \{$E_h$\} has no known relations to \{$E_t$\}                                                \\
per:religion                         & \{$E_h$\} has the religion \{$E_t$\}                                                         \\
org:country\_of\_branch              & \{$E_h$\} has a branch in the country \{$E_t$\}                                              \\
org:stateorprovince\_of\_branch      & \{$E_h$\} has a branch in the state or province \{$E_t$\}                                    \\
org:city\_of\_branch                 & \{$E_h$\} has a branch in the city \{$E_t$\}                                                 \\
org:shareholders                     & \{$E_h$\} has shares hold in \{$E_t$\}                                                       \\
org:top\_members/employees           & \{$E_h$\} has the high level member \{$E_t$\}                                                \\
org:members                          & \{$E_h$\} has the member \{$E_t$\}                                                           \\
org:website                          & \{$E_h$\} has the website \{$E_t$\}                                                          \\
per:parents                          & \{$E_h$\} has the parent \{$E_t$\}                                                           \\
org:number\_of\_employees/members    & \{$E_h$\} has the number of employees \{$E_t$\}                                              \\
org:political/religious\_affiliation & \{$E_h$\} has political affiliation with \{$E_t$\}                                           \\
per:age                              & \{$E_h$\} has the age \{$E_t$\}                                                              \\
per:origin                           & \{$E_h$\} has the nationality \{$E_t$\}                                                      \\
org:alternate\_names                 & \{$E_h$\} is also known as \{$E_t$\}                                                         \\
per:other\_family                    & \{$E_h$\} is the other family member of \{$E_t$\}                                            \\
per:identity                         & \{$E_h$\} is the identity/pronoun of \{$E_t$\} \\
per:identity                         & \{$E_h$\} and \{$E_t$\} are the same person \\
per:siblings                         & \{$E_h$\} is the siblings of \{$E_t$\}                                                       \\
org:member\_of                       & \{$E_h$\} is the member of \{$E_t$\}                                                         \\
per:children                         & \{$E_h$\} is the parent of \{$E_t$\}                                                         \\
per:employee\_of                     & \{$E_h$\} is the employee of \{$E_t$\}                                                       \\
per:spouse                           & \{$E_h$\} is the spouse of \{$E_t$\}                                                         \\
org:dissolved                        & \{$E_h$\} dissolved in \{$E_t$\}                                                             \\
per:schools\_attended                & \{$E_h$\} studied in \{$E_t$\}                                                               \\
per:country\_of\_death               & \{$E_h$\} died in the country \{$E_t$\}                                                      \\
per:stateorprovince\_of\_death       & \{$E_h$\} died in the state or province \{$E_t$\}                                            \\
per:city\_of\_death                  & \{$E_h$\} died in the city \{$E_t$\}                                                         \\
per:date\_of\_death                  & \{$E_h$\} died in the date \{$E_t$\}                                                         \\
per:cause\_of\_death                 & \{$E_h$\} died because of \{$E_t$\}                                                          \\
org:founded                          & \{$E_h$\} was founded in \{$E_t$\}                                                           \\
org:founded\_by                      & \{$E_h$\} was founded by \{$E_t$\}                                                           \\
per:countries\_of\_residence         & \{$E_h$\} lives in the country \{$E_t$\}                                                     \\
per:stateorprovinces\_of\_residence  & \{$E_h$\} lives in the state or province \{$E_t$\}                                           \\
per:cities\_of\_residence            & \{$E_h$\} lives in the city \{$E_t$\}                                                        \\
per:country\_of\_birth               & \{$E_h$\} was born in the country \{$E_t$\}                                                  \\
per:stateorprovince\_of\_birth       & \{$E_h$\} was born in the state or province \{$E_t$\}                                        \\
per:city\_of\_birth                  & \{$E_h$\} was born in the city \{$E_t$\}                                                     \\
per:date\_of\_birth                  & \{$E_h$\} has birthday on \{$E_t$\}                                                          \\
per:charges                          & \{$E_h$\} is convicted of \{$E_t$\}                                                          \\
per:title                            & \{$E_h$\} is a \{$E_t$\}                                                                     \\ \bottomrule
\end{tabular}
\caption{Templates for RETACRED datasets.}
\label{tab:appendix-sure-retacred-template}
\end{table*}
\begin{table*}[]
\centering
\small
\begin{tabular}{ll}
\toprule
\textbf{Relation}         & \textbf{Template}                          \\ \midrule
Other                     & \{subj\} has no known relations to \{obj\} \\
Component-Whole(e1,e2)    & \{subj\} is the component of \{obj\}       \\
Component-Whole(e2,e1)    & \{obj\} is the component of \{subj\}       \\
Instrument-Agency(e1,e2)  & \{subj\} is the instrument of \{obj\}      \\
Instrument-Agency(e2,e1)  & \{obj\} is the instrument of \{subj\}      \\
Member-Collection(e1,e2)  & \{subj\} is the member of \{obj\}          \\
Member-Collection(e2,e1)  & \{obj\} is the member of \{subj\}          \\
Cause-Effect(e1,e2)       & \{subj\} has the effect \{obj\}            \\
Cause-Effect(e2,e1)       & \{obj\} has the effect \{subj\}            \\
Entity-Destination(e1,e2) & \{obj\} is the destination of \{subj\}     \\
Entity-Destination(e2,e1) & \{subj\} is the destination of \{obj\}     \\
Content-Container(e1,e2)  & \{obj\} contains \{subj\}                  \\
Content-Container(e2,e1)  & \{subj\} contains \{obj\}                  \\
Message-Topic(e1,e2)      & \{obj\} is the topic of \{subj\}           \\
Message-Topic(e2,e1)      & \{subj\} is the topic of \{obj\}           \\
Product-Producer(e1,e2)   & \{obj\} produces \{subj\}                  \\
Product-Producer(e2,e1)   & \{subj\} produces \{obj\}                  \\
Entity-Origin(e1,e2)      & \{subj\} origins from \{obj\}              \\
Entity-Origin(e2,e1)      & \{obj\} origins from \{subj\}              \\ \bottomrule
\end{tabular}
\caption{Templates for SemEval datasets.}
\label{tab:appendix-sure-semeval-template}
\end{table*}


\end{document}